\documentclass[runningheads]{llncs}
\usepackage{graphicx}
\usepackage{pbox}
\usepackage{url}
\usepackage[table,xcdraw,dvipsnames]{xcolor}
\begin{document}

\title{Quality-based Pulse Estimation from NIR Face Video with Application to Driver Monitoring}
\titlerunning{Quality-based rPPG HR Estimation for Driver Monitoring}
%
\author{Javier Hernandez-Ortega\inst{1}\orcidID{0000-0001-6974-3900} \and
Shigenori Nagae\inst{2} \and
Julian Fierrez\inst{1}\orcidID{0000-0002-6343-5656} \and
Aythami Morales\inst{1}\orcidID{0000-0002-7268-4785}}
\authorrunning{J. Hernandez-Ortega et al.}
%
\institute{Universidad Autonoma de Madrid, Madrid, Spain
\email{\{javier.hernandezo,julian.fierrez,aythami.morales\}@uam.es} \and
OMRON Corporation, Kyoto, Japan
\email{shigenori.nagae@omron.com}}
\maketitle              
\begin{abstract}

In this paper we develop a robust for heart rate (HR) estimation method using face video for challenging scenarios with high variability sources such as head movement, illumination changes, vibration, blur, etc. Our method employs a quality measure $Q$ to extract a remote Plethysmography (rPPG) signal as clean as possible from a specific face video segment. Our main motivation is developing robust technology for driver monitoring. Therefore, for our experiments we use a self-collected dataset consisting of Near Infrared (NIR) videos acquired with a camera mounted in the dashboard of a real moving car. We compare the performance of a classic rPPG algorithm, and the performance of the same method, but using $Q$ for selecting which video segments present a lower amount of variability. Our results show that using the video segments with the highest quality in a realistic driving setup improves the HR estimation with a relative accuracy improvement larger than 20\%.

\keywords{Remote Plethysmography  \and Driver Monitoring \and Heart Rate \and Quality Assessment \and Face Biometrics \and NIR Video}
\end{abstract}

\section{Introduction}

\label{introduction}


Traffic accidents have become one of the main non-natural causes of death in today's society. The World Health Organization (WHO) published a report in $2018$ \cite{WHO2018} declaring that $1.35$ millions of people die annually all over the world due to traffic accidents, even becoming the main cause of death among young population (those under $30$ years old).

Some types of traffic accidents can not be predicted by any manner because they occur due to external factors such as bad weather, roads in poor condition, mechanical issues, etc. However, there is still a high amount of accidents caused by human factors that can be avoided \cite{pakgohar2011role}. For example, fatigue is one of the most common causes of accidents, and it is also one of the most preventable. Drivers experiencing fatigue have a decrease in their visual perception, reflexes, and psychomotor skills, and they may even fall asleep while driving. 

\pagebreak 

In order to reduce the number of accidents, driver monitoring has attracted a lot of research attention in the recent years \cite{lal2003development,flores2010real,awasekar975driver}. A driver monitoring system must be able to detect the presence of signals related to fatigue, allowing to take preventive actions to avoid a possible accident. Some of these actions are recommending the driver to stop in a rest area until he is fully recovered, and displaying acoustic and luminous warnings inside the car to keep the driver awake until he can stop.

Driver monitoring systems may follow different ways for achieving their target. Some of them use information about the way the driver is conducting the car, i.e. movements of the steering wheel, status of the pedals, etc \cite{Kang_2013_ICCV_Workshops}. Physiological signals such as the heart rate (HR), the blood pressure, the brain activity, etc, can also be used to detect fatigue in the driver \cite{jung2014driver}. 

A monitoring system capable of estimating physiological components such as the heart rate, or the blood pressure, may present additional benefits. These systems could be able not only to detect signs of fatigue, but also changes in the driver's general health condition. This kind of monitoring systems allow to acquire and process health information daily and non-intrusively. The captured data can be used to help doctors to make better diagnostics, or even for recommending the driver to visit a practitioner if a potential health issue is detected.

The accurate extraction of physiological signals in a real driving scenario is still a challenge. There exist different approaches depending of the acquisition method, i.e. contact-based and image-based, each one with its own strengths and weaknesses. In this paper we focus in improving the performance of an image-based method by introducing a quality assessment algorithm \cite{alonso2012Quality}. The target of this algorithm is selecting the video sequences more favorable to a specific heart rate estimation method, in a kind of quality-based processing \cite{2018_INFFUS_MCSreview2_Fierrez}.

The rest of this paper is organized as follows: Section \ref{related_works_sect} introduces driver monitoring techniques, with focus in remote photoplethysmography and its challenges. Section \ref{proposed_system} describes the proposed system.  Section \ref{databases} summarizes the dataset used. Section \ref{evaluation} describes the evaluation protocol and the results obtained. Finally, the concluding remarks and the future work are drawn in Section \ref{conclusion_section}.

\section{Driver Monitoring Techniques}
\label{related_works_sect}

\begin{table*}[t]
\caption{\textbf{Selection of works} related to pulse extraction and/or driver monitoring using contact sensors or images.}
\begin{center}
\resizebox{\textwidth}{!}{
\begin{tabular}{|c|c|c|c|c|c|}
\hline
\cellcolor[HTML]{9B9B9B}Method & \cellcolor[HTML]{9B9B9B}Type of Data & \cellcolor[HTML]{9B9B9B}Parameters Extracted & \cellcolor[HTML]{9B9B9B}Performance & \cellcolor[HTML]{9B9B9B}Target \\
\hline\hline
Brandt et al. 2004 \cite{brandt2004affordable} & RGB and NIR Video  & Head Motion and Eye Blinking & N/A & Driver Fatigue\\
Shin et al. 2010 \cite{shin2010real} & ECG  & Heart Rate & N/A & Driver Fatigue\\

Jo et al. 2011 \cite{jo2011vision} & NIR Video  & Head Pose and Eye Blinking & Accuracy = 98.55\% & Driver Drowsiness and Distraction\\
Poh et al. 2011 \cite{poh2011advancements} & RGB Video  & Heart and Breath Rate, HR Variab. & RMSE = 5.63\% & Physiological Measurement \\
Jung et al. 2014 \cite{jung2014driver} & ECG  & Heart Rate & N/A & Driver Drowsiness\\
Tasli et al. 2014 \cite{landmarks2014RGB} & RGB Video  & Heart Rate, HR Variab. & MAE = 4.2\% & Physiological Measurement\\
McDuff et al. 2014 \cite{mcduff2014multiband} & RGB-CO Video  & Heart and Breath Rate, HR Variab. & Correlation = 1.0 & Physiological Measurement\\
Chen et al. 2016 \cite{nir2016realsense} & RGB and NIR Video  & Heart Rate & RMSE = 1.65\% & Physiological Measurement \\
\hline
\textbf{Present Work} & \textbf{NIR Video} & \textbf{Heart Rate} & \textbf{MAE = 8.76\%} & \textbf{Driver Monitoring} \\
\hline
\end{tabular}
}
\end{center}
\label{related_works}
\end{table*}



Early research in driver monitoring was mostly based on acquiring accurate physiological signals from the drivers using contact sensors (e.g. ECG, EEG, or EMG), but this approach may result uncomfortable and impractical in a realistic driving environment. Some parameters that can be obtained this way are the heart rate, respiration, brain activity, muscle activation, corporal temperature, etc. Some works related to this approach are \cite{jung2014driver} and \cite{shin2010real}.

Contactless approaches are more convenient for its use in real driver monitoring without bothering the driver with cables and other uncomfortable devices. Regarding this approximation, computer vision techniques result really practical since they use images acquired non-invasively from a camera mounted inside the vehicle. These images can be processed to analyze physiological parameters using remote photoplethysmography (rPPG). With this technique it is possible to estimate the heart rate, the oxygen saturation, and other pulse related information using only video sequences \cite{poh2011advancements}. 

%


\subsection{Remote Photoplethysmography}

Photoplethysmography (PPG) \cite{allen2007photoplethysmography} is a low-cost technique for measuring the cardiovascular Blood Volume Pulse (BVP) through changes in the amount of light reflected or absorbed by human vessels. PPG is often used at hospitals to measure physiological parameters like the heart rate, the blood pressure, or  the oxygen saturation. PPG signals are usually be measured with contact sensors often placed at the fingertips, the chest, or the feet. This type of contact measurement may be suitable for a clinic environment, but it can be uncomfortable and inconvenient for daily driver monitoring. 

In recent works like \cite{poh2011advancements}, \cite{landmarks2014RGB}, \cite{mcduff2014multiband}, and \cite{nir2016realsense} remote photoplethysmography techniques have been used for measuring physiological signals from face video sequences captured at distance. These works used signal processing techniques for analyzing the images, and looking for slight color and illumination changes related with the BVP. However, using these methods in a real moving vehicle is not straightforward due to all the variability sources present in this type of video sequences. A selection of works related to driver monitoring and photoplethysmography is shown in Table \ref{related_works}.

\subsection{Challenges and Proposed Approach}

A moving vehicle is not a perfect environment for obtaining high accuracy when using rPPG algorithms. Images acquired in this scenario may present external illumination changes, low illumination levels, noise, movement of the driver, occlusions, and vibrations of the camera due to the movement of the vehicle. All these factors can make the performance of the rPPG algorithms to drop significantly \cite{alonso2012Quality}.

In this work we propose a system for pulse estimation for driver monitoring that tries to overcome some of these challenges. We use a NIR camera with active infrared illumination mounted in the dashboard of a real moving car. The NIR spectrum band is highly invariant to ambient light, providing robustness against this external source of variability at a low cost. This also allowed us to extend the application of heart rate estimation to very low illumination environments, e.g. night conditions.

Regarding to the presence of other variability factors such as movement or occlusions, a quality-based approach to rPPG could be adequate \cite{2018_INFFUS_MCSreview2_Fierrez}. With a short-time analysis, small video segments without enough quality for extracting a robust rPPG signal could be discarded without affecting the global performance of pulse estimation. To accomplish this target, we have proposed a quality metric for short segments of rPPG signals.

Summarizing, in this work: i) we performed pulse estimation using NIR active illumination to be robust to external illumination variability; ii) we proposed a quality metric for classifying short rPPG segments and deciding which ones can be used and which ones should be discarded in order to obtain a robust heart rate estimation; and iii) we compared the performance of a classic rPPG algorithm and our quality-based approach.

\section{Proposed System}
\label{proposed_system}

In this section, we describe the improvements we have done to a baseline rPPG-based heart rate estimation system to increase its performance in a real driving scenario. Classic rPPG systems drastically degrade when facing the variability sources mentioned in the previous sections. This performance problem is caused by the low quality of the extracted rPPG signals which may be affected (in their totality or only in some fragments) by variability sources that the rPPG method does not know how to deal with.

Having this into mind, we thought that computing a quality measure for knowing the amount of variability in each temporal segment of a rPPG signal could be useful for deciding which segments are more suitable for extracting a robust heart rate estimation.

\begin{figure*}[t!]
\begin{center}
   \includegraphics[width=\linewidth]{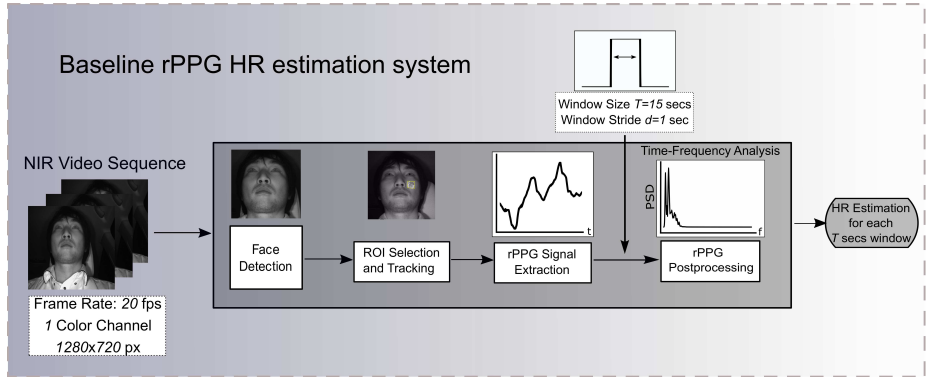}
\end{center}
   \caption{\textbf{Architecture of the baseline rPPG system for HR estimation}. Given a facial NIR video, the face is detected and the rPPG signal is extracted from the ROI. The raw rPPG signal is windowed and postprocessed in order to obtain an individual HR estimation for each video segment.}
\label{baseline_scheme}
\end{figure*}

In the next subsection we describe the vanilla rPPG system we used to obtain the baseline results. This method corresponds to the system shown in Figure \ref{baseline_scheme}. In the second subsection we describe the addition of a quality metric to the baseline system. That approach is shown in Figure \ref{quality_scheme}. In the third subsection we describe how we have obtained the groundtruth of the heart rate for our experiments.

\subsection{Baseline rPPG System}

The basic method is based in the one used in \cite{CVPRw2018}, and consists of the next three main steps:

\begin{itemize}

\item \textbf{Face detection and ROI tracking:} The first step consists in detecting the face of the driver on the first frame of the NIR video. We used the Matlab implementation of the Viola-Jones algorithm. This algorithm is known to perform reasonably well and in real time when dealing with frontal faces, as in our case.  After the recognition stage we selected the left cheek as the Region Of Interest (ROI), since it is a zone lowly affected by objects like hats, glasses, beards, or mustaches. The next step consisted of detecting corners inside the ROI for tracking them over time using the Kanade-Lucas-Tomasi algorithm, also implemented in Matlab. If at some point of the video the ROI is lost, the face will be redetected, and after that also the ROI and the corners.\\

\item \textbf{rPPG signal extraction:} For each frame from the video, we calculated its raw rPPG value as the averaged intensity of the pixels inside the ROI. The final output for each video sequence is a rPPG temporal signal composed by the concatenation of these averaged intensities.\\

\item \textbf{rPPG postprocessing:} We wanted to estimate a HR value each $d$ seconds. In order to achieve that target, we extracted windows of $T$ seconds from the rPPG signal, with a stride of $d$ seconds between them. The length of the window ($T$) is configurable in order to perform a time dependent analysis. For each window we postprocessed the raw rPPG signal and we obtained an estimation of the HR. This postprocessing method consists of three filters:\\

\begin{itemize}
\item Detrending filter: this temporal filter is employed for reducing the stationary part of the rPPG signal, i.e. eliminating the contribution from environmental light and reducing the slow changes in the rPPG level that are not part of the expected pulse signal.\\

\item Moving-average filter: this filter is designed to eliminate the random noise on the rPPG signal. That noise may be caused by imperfections on the sensor and inaccuracies in the capturing process. This filter consists in a moving average of the rPPG values (size $3$).\\

\item Band-pass filter: we considered that a regular human heart rate uses to be into the $40$-$180$ beats per minute (bpm) range, which corresponds to signals with frequencies between $0.7$ Hz and $3$ Hz approximately. All the rPPG frequency components outside that range are unlikely to correspond to the real pulse signal so they are discarded.\\
\end{itemize}

After this processing stage we transformed the signal from the time domain to the frequency domain using the Fast Fourier Transform (FFT). Then, we estimated its Power Spectral Density (PSD) distribution. Finally, we searched for the maximum value in that PSD. The frequency correspondent to that maximum is the estimated HR of that specific video segment.

\end{itemize}

\subsection{Proposed Quality-Based Approach}	

\begin{figure*}[t!]
\begin{center}
   \includegraphics[width=\linewidth]{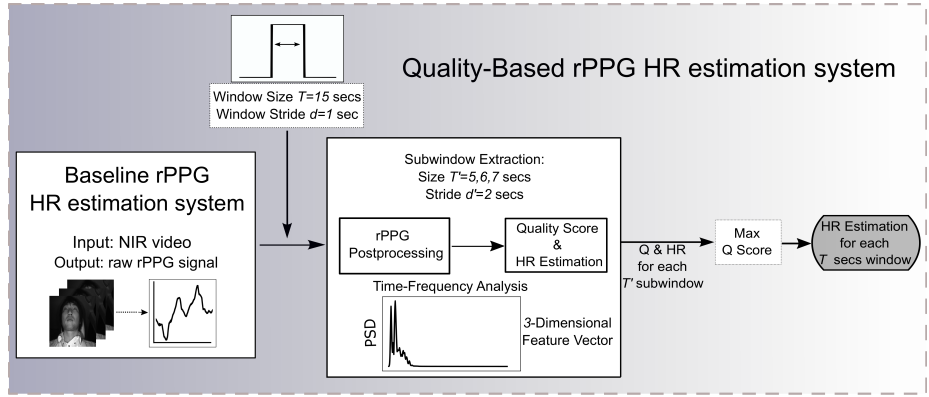}
\end{center}
   \caption{\textbf{Architecture of the quality-based rPPG system for HR estimation}. We extracted some features from subwindows of the postprocessed rPPG signal. These features were used to compute a quality metric for estimating the presence of noise, head motion, or external illumination variability in the rPPG signal. For each $T$ seconds window we selected the $T'$ seconds subwindow with the highest quality.}
\label{quality_scheme}
\end{figure*}

The baseline method is able to obtain robust HR estimations in controlled scenarios without too much variability or noise in the recordings. However, the raw rPPG signals acquired in a realistic driver monitoring scenario use to have high variations due to external illumination changes, and frequent movements of the driver’s head. There are also other sources of noise, e.g. noise inherent to the acquisition sensor. 

All the mentioned factors make the performance of the baseline rPPG algorithm to dramatically fall. In order to make it as robust as possible, we decided to develop an new approach, consisting of an evolution of the basic system combined with a quality metric of the raw rPPG signals. A scheme of the proposed quality-based method can be seen in Figure \ref{quality_scheme}.

The target of using the quality metric is selecting the temporal subwindow of $T’$ seconds with the highest quality from all the subwindows available inside each $T$ seconds window. The criteria for determining the best quality consists in looking for the rPPG segment with the less presence of noise, head motion, and external illumination variability, i.e. the rPPG signal closest to one that has been captured with a contact sensor.
 
In order to compute the quality level, we divided each window into several subwindows of $T’$ seconds, with a stride of $d’$ seconds between them (both parameters are configurable). Then we performed the processing of the rPPG signal in the same way done in the baseline system. From each processed rPPG subwindow we extracted several features, and we combined them to obtain a single numerical quality measure ($Q$) representative of how close is the rPPG signal of each subwindow to one acquired in perfect conditions. 

Finally, from each $T$ seconds window, we selected the segment of $T’$ seconds with the highest $Q$, and we estimate the user's HR with that rPPG segment. This way we discarded the rPPG fragments that may be more affected by variability. This value of the HR is used as the final HR estimation for the whole $T$ seconds window.

\section{Dataset}
\label{databases}

\subsection{OMRON Database}

\begin{figure*}[t!]
\begin{center}
   \includegraphics[width=\linewidth]{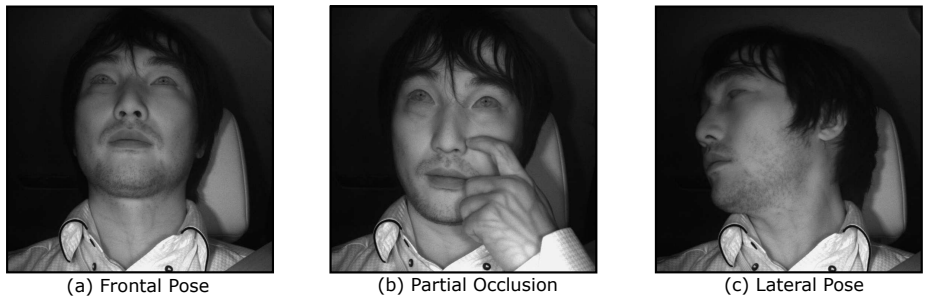}
\end{center}
   \caption{\textbf{Images extracted from the OMRON database}: (a) shows a image with a low level of variability. A high quality rPPG signal could be extracted from a video composed by this type of images; (b) and (c) show examples
of images with high variability, such as occlusions and head rotation respectively.}
\label{database}
\end{figure*}

We tested our method with a self-collected dataset called OMRON Database. The data in the dataset is composed by Near Infrared (NIR) active videos of the driver's faces, recorded with a camera mounted in a car dashboard. The images were captured at a sampling rate of $20$ fps, and a resolution of $1280\times720$ pixels. The PPG signals used for the groundtruth were captured using a BVP fingerclip sensor with a sampling rate of $500$ Hz, and then downsampled to $20$ Hz to synchronize them with the images from the camera.

The dataset is comprised of $7$ male users, with different ages, skin tones and some of them wearing glasses. Each participant was in front of the camera during a single session with a different duration for each one. The sessions went from $20$ minutes to $60$ minutes long. The full database contains $400,000$ images with an average of $57,000$ images for each subject. The recordings try to represent a real driving scenario inside a moving car. They present different types of variability such as head movement, occlusions, car vibration, or external illumination. These variations mean different levels of quality in the estimated rPPG signals. Examples of images from this database can be seen in Figure \ref{database}.

\section{Evaluation}
\label{evaluation}

In this section we compare the performance of the heart rate estimations obtained using the quality-based rPPG method with the performance obtained using the baseline rPPG method.

%

\subsection{Setting quality parameters and features}

\begin{table}[t]
\caption{ \textbf{Left: Features extracted to compute the quality} of the rPPG postprocessed signals. \textbf{Right: Final configuration of the parameters} of the quality-based method. \emph{*From each window of $T$ seconds, we extracted subwindows of $T' =$  $5$, $6$, and $7$ seconds of duration, and we selected the one with the highest $Q$ value.}}
\resizebox{\textwidth}{!}{
\begin{tabular}{ccl|c|c|}
\cline{1-2} \cline{4-5}
\multicolumn{1}{|c|}{\cellcolor[HTML]{9B9B9B}\textbf{Feature}} & \multicolumn{1}{c|}{\cellcolor[HTML]{9B9B9B}\textbf{Description}}                                                                                                     &  & \cellcolor[HTML]{9B9B9B}\textbf{Parameter} & \cellcolor[HTML]{9B9B9B}\textbf{Value} \\ \cline{1-2} \cline{4-5} 
\multicolumn{1}{|c|}{Signal Noise Ratio (SNR)}                 & \multicolumn{1}{c|}{\begin{tabular}[c]{@{}c@{}}Power of the maximum value in the PSD\\ and its two first harmonics,\\ divided by the rest of the power.\end{tabular}} &  & Window Size T                              & 7 seconds                              \\ \cline{1-2} \cline{4-5} 
\multicolumn{1}{|c|}{Bandwidth (BW)}                           & \multicolumn{1}{c|}{\begin{tabular}[c]{@{}c@{}}Bandwidth containing the 99\% of the power,\\ centered in the maximum value of the PSD.\end{tabular}}                  &  & Window Stride d                            & 1 second                               \\ \cline{1-2} \cline{4-5} 
\multicolumn{1}{|c|}{Ratio Peaks (RP)}                         & \multicolumn{1}{c|}{\begin{tabular}[c]{@{}c@{}}Power of the highest peak in the PSD divided\\ by the power of second highest peak.\end{tabular}}                      &  & Subwindow Size T'                         & 5, 6, and 7 seconds*                   \\ \cline{1-2} \cline{4-5} 
\multicolumn{1}{l}{}                                           & \multicolumn{1}{l}{}                                                                                                                                                  &  & Subwindow Stride d'                        & 2 seconds                              \\ \cline{4-5} 
\multicolumn{1}{l}{}                                           & \multicolumn{1}{l}{}                                                                                                                                                  &  & Feature Vector                             & SNR, BW, and RP                        \\ \cline{4-5} 
\end{tabular}
}
\label{optim_config}
\end{table}


The quality-based method has several parameters to be configured: the window size $T$, the subwindow size $T’$, the window stride $d$, and the subwindow stride $d’$. It is also necessary to decide which features to extract from the rPPG signals, as they must contain information about the quality level of each subwindow $T'$.

For this work we extracted $3$ different features that can give us information about how close/far is a rPPG signal from the one captured in perfect conditions. The features and their descriptions can be seen in Table~\ref{optim_config} (left). The final quality metric is computed as the arithmetic mean of these $3$ features after normalizing them to the [$0$,$1$] range using a $tanh$ normalization \cite{2018_INFFUS_MCSreview2_Fierrez}.

Based on our own previous rPPG experiments, we decided to test values of $T$ going from $5$ seconds to $15$ seconds, with $1$ second of increment for the loop. From our previous work \cite{CVPRw2018} we know that $T=5$ seconds was the lowest value that gave good HR estimation with favorable conditions, and using windows longer than $15$ seconds did not show to improve the results.

For setting the subwindow duration $T’$, we decided to test values going from $5$ seconds (limited by the minimum possible $T$ size), to the correspondent $T$ value in each case. We also incremented the $T'$ values using a step of $1$ second. The stride $d$ is set to $1$ second in order to give an estimation of the HR for each second of the input video. The stride $d’$, i.e. the temporal step between each subwindow, took values going from a minimum of $1$ second to a maximum of $5$ seconds (when possible), with $1$ second of increment. After this initial configuration experiments the best results were obtained for the parameters shown in Table \ref{optim_config} (right).



To compute the performance of heart rate estimations we decided to use the Mean Absolute Error (MAE) between the groundtruth heart rate in beats per minute (bpm), and the one estimated with the rPPG algorithm.


\subsection{Results}

For the final evaluation of both methods (baseline and proposed quality-based), we processed $19$ NIR videos of $1$ minute duration each one from the OMRON Database. We used the configuration of parameters shown in Table \ref{optim_config} (right) for both methods (only $T$ and $d$ in the case of the baseline system). We first computed the Mean Absolute Error (MAE) for each NIR video separately. We did this to have an idea of which videos are working better and which ones are working worse. We also computed the mean and standard deviation of the MAE for the whole evaluation dataset.

\begin{table*}[t]
\caption{Results of HR estimation for the rPPG baseline method and the quality-based approach. The results comprehend the individual Mean Absolute Errors (MAE) with respect to the groundtruth HR. The mean value and the standard deviation of the MAE for the whole evaluation data has been also computed. The relative improvement of the MAE is shown between parentheses.
}
\begin{center}
\resizebox{\textwidth}{!}{
\begin{tabular}{cc|c|c|c|c|c|c|c|c|c|c|c|}
\hline
\multicolumn{1}{|c|}{\cellcolor[HTML]{FFFFFF}\textbf{MAE [bpm]}} & \cellcolor[HTML]{9B9B9B}Video Number & \cellcolor[HTML]{9B9B9B}1 & \cellcolor[HTML]{9B9B9B}2 & \cellcolor[HTML]{9B9B9B}3 & \cellcolor[HTML]{9B9B9B}4 & \cellcolor[HTML]{9B9B9B}5 & \cellcolor[HTML]{9B9B9B}6 & \cellcolor[HTML]{9B9B9B}7 & \cellcolor[HTML]{9B9B9B}8 & \cellcolor[HTML]{9B9B9B}9 & \cellcolor[HTML]{9B9B9B}10 & \cellcolor[HTML]{9B9B9B}11\\
\hline

\multicolumn{2}{|c|}{\cellcolor[HTML]{DDDDDD}Baseline Method} &11.9	&15.9	&7.5	&12.6	&11.2	&12.5	&9.6	&\cellcolor[HTML]{009B00}8.0	&9.2 &9.6	&\cellcolor[HTML]{009B00}10.1\\
\hline

\multicolumn{2}{|c|}{\cellcolor[HTML]{DDDDDD}Proposed Method} & \cellcolor[HTML]{009B00}9.1	&\cellcolor[HTML]{009B00}9.9	&\cellcolor[HTML]{009B00}7.5	&\cellcolor[HTML]{009B00}8.1	&\cellcolor[HTML]{009B00}10.9	&\cellcolor[HTML]{009B00}8.2	&\cellcolor[HTML]{009B00}7.9	&8.7	&\cellcolor[HTML]{009B00}5.7 & \cellcolor[HTML]{009B00}5.8	&10.5 \\
\hline

 \multicolumn{1}{c}{}& \multicolumn{1}{c}{}& \multicolumn{1}{c}{} &\multicolumn{1}{c}{}  & \multicolumn{1}{c}{} & \multicolumn{1}{c}{} & \multicolumn{1}{c}{} & \multicolumn{1}{c}{} & \multicolumn{1}{c}{} &\multicolumn{1}{c}{} & \multicolumn{1}{c}{} & \multicolumn{1}{c}{}\\

\hline
\multicolumn{1}{|c|}{\cellcolor[HTML]{FFFFFF}\textbf{MAE [bpm]}}& \cellcolor[HTML]{9B9B9B}Video Number  & \cellcolor[HTML]{9B9B9B}12 & \cellcolor[HTML]{9B9B9B}13 & \cellcolor[HTML]{9B9B9B}14& \cellcolor[HTML]{9B9B9B}15 & \cellcolor[HTML]{9B9B9B}16 & \cellcolor[HTML]{9B9B9B}17 & \cellcolor[HTML]{9B9B9B}18 & \cellcolor[HTML]{9B9B9B}19 &  \cellcolor[HTML]{9B9B9B}Mean & \multicolumn{1}{c|}{\cellcolor[HTML]{9B9B9B}Std} & \multicolumn{1}{c|}{\cellcolor[HTML]{9B9B9B}}\\ 
\hline

\multicolumn{2}{|c|}{\cellcolor[HTML]{DDDDDD}Baseline Method} & \cellcolor[HTML]{009B00}8.8	&14.6	&10.3	&13.1	&\cellcolor[HTML]{009B00}7.9	&11.1	&\cellcolor[HTML]{009B00}9.8 &15.2	&\cellcolor[HTML]{FFFFFF}11.0	&\cellcolor[HTML]{FFFFFF}2.4 & \multicolumn{1}{c|}{\cellcolor[HTML]{9B9B9B}}\\
\hline

\multicolumn{2}{|c|}{\cellcolor[HTML]{DDDDDD}Proposed Method} & 9.3	&\cellcolor[HTML]{009B00}7.6	&\cellcolor[HTML]{009B00}7.0	&\cellcolor[HTML]{009B00}7.9	&8.4	&\cellcolor[HTML]{009B00}9.1	&11.2	&\cellcolor[HTML]{009B00}12.7	&\cellcolor[HTML]{FFFFFF}\begin{tabular}[c]{@{}c@{}}8.7\\ (\textcolor{OliveGreen}{\textbf{21\%}})\end{tabular}	& \cellcolor[HTML]{FFFFFF}\begin{tabular}[c]{@{}c@{}}1.7\\ (\textcolor{OliveGreen}{\textbf{29\%}})\end{tabular} & \multicolumn{1}{c|}{\cellcolor[HTML]{9B9B9B}}\\
\hline

\end{tabular}
}
\end{center}
\label{result_comparison}
\end{table*}

\begin{figure*}[t]
\begin{center}
   \includegraphics[width=0.8\linewidth]{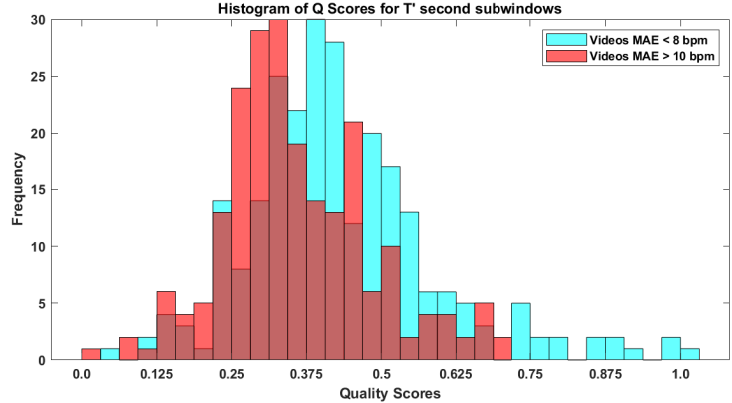}
\end{center}
   \caption{\textbf{Quality Scores obtained from $T'$ seconds windows.} We have selected those videos with a mean MAE under $8$ bpm as representative of high quality videos, and those with a mean MAE over $10$ bpm as low quality videos. The histograms show two different distributions, with the high quality videos presenting a higher mean value of the quality score $Q$.}
\label{Q_scores_histogram}
\end{figure*}

As can be seen in Table \ref{result_comparison}, using the quality-based rPPG approach we obtained a MAE value averaged across videos of $8.7$ beats per minute (bpm), and a standard deviation of $1.7$ bpm for the whole evaluation dataset. Compared to this result, the baseline system (without the quality approach) obtained a MAE of $11.0$ bpm and a MAE standard deviation of $2.4$ bpm for the $19$ minutes. This difference in the performance represents a relative improvement of a $21$\% in the mean value, and of the $29$\% in the standard deviation of the Mean Absolute Error.

Table \ref{result_comparison} also shows the MAE values for each NIR video of the evaluation dataset, both using the baseline and the quality-based methods. It can be seen that for some specific videos the baseline result is obtaining a more accurate estimation of the HR, but in general, the MAE values obtained using the quality-based approach are lower. The specific cases in which the quality-based method is working worse coincide with those videos with long sequences with high variability, what makes difficult to find clean segments.

In Figure \ref{Q_scores_histogram} we are showing the quality scores $Q$ we obtained for a selection of the evaluation videos. We decided to show the distribution of $Q$ scores from those videos with a MAE value (obtained with the quality-based method) lower than $8$ bpm, and those with a MAE value higher than $10$ bpm. The histograms show two different distributions, with the best performing videos (i.e., MAE < 8 bpm) presenting a higher mean value of the quality score $Q$. 


The results of this section evidenced that, at least with the data from the OMRON Dataset, the quality metric $Q$ has shown to be an effective way to discard segments of video that may impact negatively to the general performance in rPPG, and therefore obtaining an improvement of the global accuracy of HR estimation.

\section{Conclusion and Future Work}
\label{conclusion_section}

In this paper we developed a method for improving heart rate (HR) estimation using remote photoplethysmography (rPPG) in challenging scenarios with multiple sources of high-variability and degradation. Our method employs a quality measure to extract a rPPG signal as clean as possible from a specific face video segment, trying to obtain a more robust HR estimation.

Our main motivation is developing robust technology for contactless driver monitoring using computer vision. Therefore, in our experiments we employed Near Infrared (NIR) videos acquired with a camera mounted in a car dashboard. This type of videos present a high number of variability sources such as head movement, external illumination changes, vibration, blur, etc. The target of the quality metric $Q$ we have proposed consists in estimating the amount of presence of those factors. Even though our experimental framework is around driver monitoring, our methods may find application in other high-variability face-based human-computer interaction scenarios such as mobile video-chat.

We have compared the performance of two different methods for HR estimation using rPPG. The first one consisted in a classic rPPG algorithm. The second method consisted in the same algorithm, but using the quality measure $Q$ for selecting which video segments present a lower amount of variability. We used those segments for extracting rPPG signals and their associated HR estimations. The quality metric $Q$ showed to be a reliable estimation of the amount of variability. We achieved better performance in HR estimation using the video segments with the highest possible quality, compared to using all the video frames indistinctly.

Our solution is based on defining the quality $Q$ as a combination of hand-crafted features. As future work, other definitions of quality could be also investigated. A different set of features that may correlate more accurately to the presence of noise factors in the rPPG signal can be studied. Training a Deep Neural Network (DNN) for extracting $Q$ from the video sequences is also an interesting possibility. This type of networks may be able of estimating the quality level by learning which factors are more relevant for obtaining robust rPPG signals directly from training data. However, the lack of labeled datasets makes it difficult to train DNNs from scratch, so it would be also beneficial to acquire a larger database. This new database may contain a higher number of users, and it may also present more challenging conditions for testing our quality-based rPPG algorithm, e.g. variant ambient illumination, motion, blur, occlusions, etc.


\section{Acknowledgements}
This work was supported in part by projects BIBECA (RTI2018-101248-B-I00 from MICINN/FEDER), and BioGuard (Ayudas Fundacion BBVA). The work was conducted in part during a research stay of J. H.-O. at the Vision Sensing Laboratory, Sensing Technology Research Center, Technology and Intellectual Property H.Q.,OMRON Corporation, Kyoto, Japan. He is also supported by a PhD Scholarship from UAM.

 \bibliographystyle{splncs04}
 \bibliography{egbib}

\end{document}